\documentclass[journal]{IEEEtran}
\usepackage[utf8]{inputenc}
\usepackage{cite}
\usepackage{hyperref}
\usepackage{multirow}
\usepackage{algorithmic}
\usepackage{orcidlink}
\usepackage{subfigure}
\usepackage{amsmath} 
\usepackage{amsfonts,amssymb}  
\usepackage{colortbl}
\usepackage{algorithm,algorithmic}
\usepackage{graphicx} 
\usepackage{titlesec} 
\titlespacing*{\section}{0pt}{0.3em}{0.3em}  
\titlespacing*{\subsection}{0pt}{0.2em}{0.2em}  
\graphicspath{ {./figures/} }
\renewcommand\eqref[1]{(\autoref{#1})}
\hypersetup{
	colorlinks=true,
	linkcolor=cyan,
	filecolor=blue,      
	urlcolor=black,
	citecolor=green,
}

\begin{document}

\title{\textit{m}HC-HSI: Clustering-Guided Hyper-Connection Mamba for Hyperspectral Image Classification}

\author{
Yimin Zhu,  Zack Dewis, Quinn Ledingham, Saeid Taleghanidoozdoozan, Mabel Heffring, Zhengsen Xu,  Motasem Alkayid,  Megan Greenwood, Lincoln Linlin Xu, ~\IEEEmembership{Member,~IEEE} 


\thanks{Authors are all from the Department of Geomatics Engineering, University of Calgary, Canada,  email: yimin.zhu@ucalgary.ca, Corresponding author: Lincoln Linlin Xu, email: lincoln.xu@ucalgary.ca. Motasem Alkayid is also affiliated with the Department of Geogra-
phy, Faculty of Arts, The University of Jordan, Amman, Jordan}}

\markboth{Journal of \LaTeX\ Class Files,~Vol.~13, No.~9, September~2014}
{Shell \MakeLowercase{\textit{et al.}}: }
\maketitle

\begin{abstract}

Recently, DeepSeek has invented the manifold-constrained hyper-connection (\textit{m}HC) approach which has demonstrated significant improvements over the traditional residual connection in deep learning models \cite{xie2026mhc}. Nevertheless, this approach has not been tailor-designed for improving hyperspectral image (HSI) classification. This paper presents a clustering-guided \textit{m}HC Mamba model (\textit{m}HC-HSI) for enhanced HSI classification, with the following contributions. First, to improve spatial-spectral feature learning, we design a novel clustering-guided Mamba module, based on the \textit{m}HC framework, that explicitly learns both spatial and spectral information in HSI. Second, to decompose the complex and heterogeneous HSI into smaller clusters, we design a new implementation of the residual matrix in mHC, which can be treated as soft cluster membership maps, leading to improved explainability of the \textit{m}HC approach. Third, to leverage the physical spectral knowledge, we divide the spectral bands into physically-meaningful groups and use them as the "parallel streams" in \textit{m}HC, leading to a physically-meaningful approach with enhanced interpretability. The proposed approach is tested on benchmark datasets in comparison with the state-of-the-art methods, and the results suggest that the proposed model not only improves the accuracy but also enhances the model explainability. Code is available here: \url{https://github.com/GSIL-UCalgary/mHC_HyperSpectral}.

\end{abstract}


\IEEEpeerreviewmaketitle

\section{Introduction}

\IEEEPARstart{H}{yperspectral} image (HSI) classification is a fundamental task that transforms raw HSI data into valuable maps that support various critical environmental and resource exploitation
tasks. Nevertheless, HSI classification is a challenging task due to the complex spatial-spectral heterogenous patterns, leading to difficulties in extracting discriminative features and also challenges in explaining the model performance \cite{SSRN,9565208}. Therefore, it is critical to develop models that are excellent not only in improving the classification accuracy but also in enhancing the model explainability.  

The convolutional neural network (CNN) approaches have been widely used for improving HSI classification performance by enabling better feature learning capability than traditional machine learning approaches \cite{SSRN}. Comparing with the CNN approaches, the Transformers models are better at learning the long-distance large-scale spatial-spectral context dependency \cite{9565208}. However, Transformer's self-attention mechanism has very high computational complexity which grows quadratically with respect to the size of the image (or sequence length). The Mamba model, which can also model the long-range context information, overcomes the drawbacks of Transformers by adopting state recursion
and sequential tokens, leading to linear complexity. 

\begin{figure}[!t]
    \centering
    \includegraphics[width=0.5\textwidth]{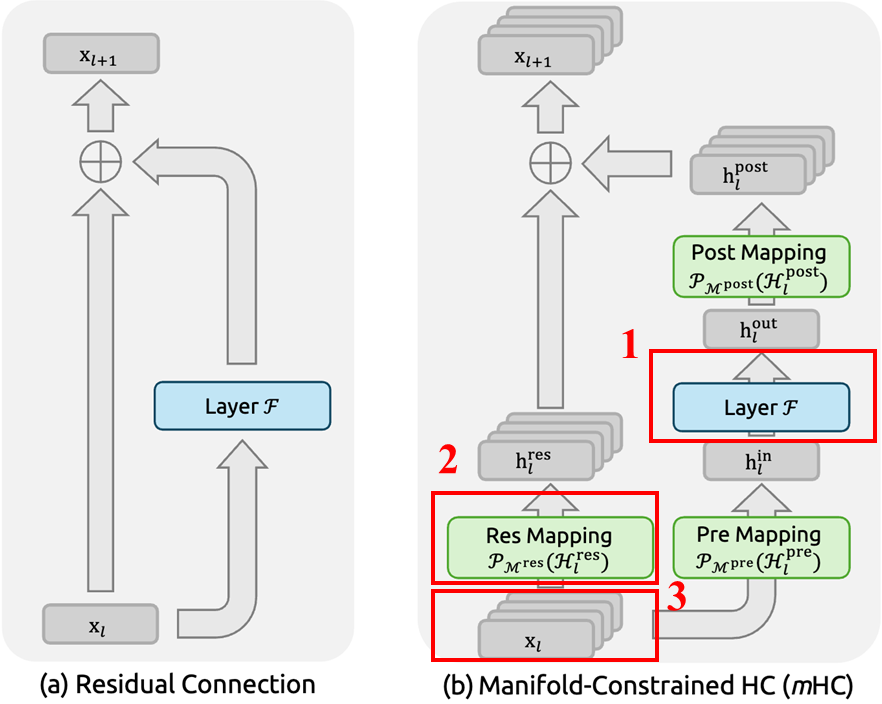}
    \caption{Figure from \cite{xie2026mhc}. Comparison between (a) a standard residual connection and (b) the Manifold-Constrained Hyper-Connections (\textit{m}HC) framework. The three red bounding boxes in (b) highlight the key components where our contributions are introduced. Specifically, they correspond to: (1) a clustering-guided Mamba module for enhanced spatial–spectral feature learning in hyperspectral images, (2) a novel residual mapping implementation that can be interpreted as soft cluster membership maps to decompose heterogeneous scenes and improve model explainability, and (3) a physically meaningful multi-stream design based on spectral band grouping to incorporate spectral prior knowledge. Together, these modifications adapt the original \textit{m}HC framework for hyperspectral image classification with improved accuracy and interpretability. }
    \label{deepseek_mHC}
\end{figure}

However, the traditional vision Mamba models \cite{zhu2024vision} treat the entire image, which has complex spatial-spectral patterns, as one token sequence, and build a very long global token sequence to address this image, leading to not only high computational cost, but also the notorious correlation decay problem that bottlenecks Mamba performance. Therefore, how to break down the complex image into simpler clusters and improve Mamba using the clustering-guided approach is an important research issue.

Residual connections are fundamental in modern AI for overcoming gradient decay and have significantly advanced HSI classification~\cite{SSRN}. However, their fixed addition creates an information bottleneck, limiting selective feature propagation across depths~\cite{zhu2024hyper}. Hyper-Connections (HC) address this using learnable matrices between residual streams, but their unconstrained nature compromises the identity mapping property, leading to gradient explosion~\cite{zhu2024hyper, xie2026mhc}. 

Recently, the Manifold-Constrained Hyper-Connections (\textit{m}HC) approach~\cite{xie2026mhc} resolves this fundamental tension between enhanced connectivity and training stability, as illustrated in \autoref{deepseek_mHC} (b). The solution has two key components: first, it preserves enhanced representational capacity by replicating the input \(\mathbf{x}_l\) into a multi-lane stream, where a \textit{Pre Mapping} \(\mathcal{H}^{\text{pre}}_l\) aggregates features for the core function \(\mathcal{F}(\cdot)\) with parameter \(\mathcal{W}\) and a \textit{Post Mapping} \(\mathcal{H}^{\text{post}}_l\) expands the output back. Second, and most critically, the \textit{Res Mapping} matrix \(\mathcal{H}^{\text{res}}\) is projected onto the Birkhoff polytope of doubly stochastic matrices via Sinkhorn-Knopp normalization to become \(\mathcal{H}^{\text{res}}_{l\mathcal{M}}\). This manifold constraint ensures the composite mapping across layers preserves the feature mean and regularizes signal norm, thereby restoring the stable identity mapping property that HC lacked. The layer's operation,
\begin{align}
    \begin{aligned}
        \mathbf{x}_{l+1} = \mathcal{H}^{\text{res}}_{l\mathcal{M}}\mathbf{x}_{l} + \mathcal{H}^{\text{post}\top}_{l}\mathcal{F}(\mathcal{H}^{\text{pre}}_{l}\mathbf{x}_{l},\mathcal{W}_{l})
    \end{aligned}
\end{align}
thus provides greater representational capacity than the standard residual connection \(\mathbf{x}_{l+1} = \mathbf{x}_l + \mathcal{F}(\mathbf{x}_l)\) while preventing the signal instability of unconstrained hyper-connections.

Although this \textit{m}HC approach has demonstrated improved performance comparing with the traditional residual connection approach, it has not been tailor designed to improve HSI classification. Moreover, how to integrate \textit{m}HC with the clustering-guided Mamba approach to enhance both HSI classification accuracy and model explainability is an important research issue. 

This paper therefore presents a clustering-guided \textit{m}HC Mamba (\textit{m}HC-HSI) for enhanced HSI classification, with the following contributions. As shown by the red bounding boxes in \autoref{deepseek_mHC}.

(1) first, to improve spatial-spectral feature learning, based on the \textit{m}HC framework, we implement \(F(\cdot)\) with a novel clustering-guided Mamba module that explicitly learns both the spatial and spectral information in HSI.

(2) Second, to decompose the complex and heterogeneous HSI into smaller clusters, we design a new implementation of the residual matrix \(\mathcal{H}^{\text{res}}_{l\mathcal{M}}\) in \textit{m}HC, which can be treated as soft cluster membership maps, leading to improved explainability of the \textit{m}HC approach. 

(3) Third, to leverage the physical spectral knowledge, we divide the input hyperspectral bands into physically-meaningful groups and use them as the "streams" in  \(x_l\), leading to a physically-meaningful approach with enhanced interpretability.




\begin{figure*}[!h]
    \centering
    \includegraphics[width=\textwidth]{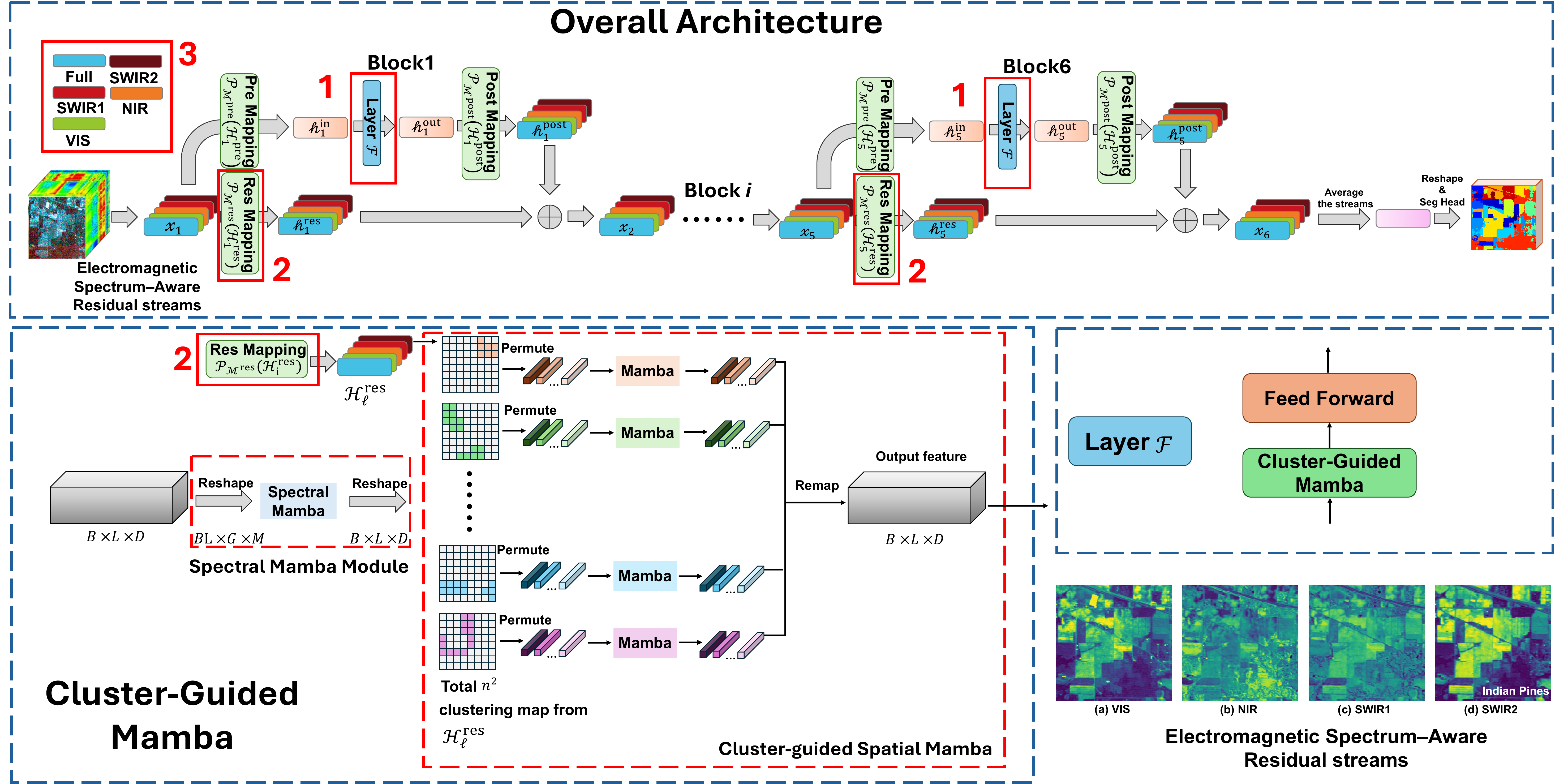}
    \caption{The architecture of the proposed clustering-guided \textit{m}HC Mamba (\textit{m}HC-HSI) model. Based on the \textit{m}HC framework, we implement \(F(\cdot)\) with a novel clustering-guided Mamba module, which has a spectral Mamba module followed by the clustering-guided spatial Mamba module. Second, we implement the residual matrix \(\mathcal{H}^{\text{res}}\) as soft cluster membership maps, which are used to decompose the complex and heterogeneous HSI into smaller clusters in the clustering-guided spatial Mamba. Third, we divide the input the spectral bands in the input HSI into physically-meaningful groups and use them as the "streams" in \(x_l\).}
    \label{fig1}
\end{figure*}

\section{Methodology} \label{method}

\subsection{Overview} \label{overall}
As shown in \autoref{fig1}, there are six residual blocks, where each block has two parallel paths, i.e., residual path with stream interaction (lower path) and feature extraction (upper path). For stream feature \(\boldsymbol{R}\) at block \(l\), overall feature update as follows:
\begin{align}
    \begin{aligned}
        \boldsymbol{R}_{l+1} = \mathcal{H}^{\text{res}}_{l\mathcal{M}}\boldsymbol{R}_{l} + \mathcal{H}^{\text{post}\top}_{l}\mathcal{F}(\mathcal{H}^{\text{pre}}_{l}\boldsymbol{R}_{l},\mathcal{W}_{l})
    \end{aligned}
\end{align}
where \(\mathcal{H}^{\text{res}}_{l\mathcal{M}}\boldsymbol{R}_{l}\) is residual path with stream interaction branch and \(\mathcal{H}^{\text{post}\top}_{l}\mathcal{F}(\mathcal{H}^{\text{pre}}_{l}\boldsymbol{R}_{l},\mathcal{W}_{l})\) represents feature extraction layer. Function \(\mathcal{F}\) is implemented by cluster-guided Mamba. \(\mathcal{H}^{\text{pre}} \in \mathbb{R}^{L \times 1 \times n}\) is stream aggregation matrix,  \(\mathcal{H}^{\text{post}} \in \mathbb{R}^{L \times 1 \times n}\) is stream expansion matrix, and \( \mathcal{H}^{\text{res}}_{l\mathcal{M}} \in \mathbb{R}^{L \times n \times n}\) is the residual matrix, where $L=H \times W$. All three key matrices, i.e., \(\mathcal{H}^{\text{res}}\), \(\mathcal{H}^{\text{pre}}\) and \(\mathcal{H}^{\text{post}}\) are  learned from the features. Algorithm \autoref{algo1} shows more details, in which CGM represents the clustering-Guided Mamba. The three key contributions are illustrated below.

\begin{itemize}
    \item \textbf{Clustering-Guided Mamba}: In \autoref{fig1}, the feature layer function \(\mathcal{F(.)}\), highlighted by red bounding box 1, is achieved via a novel clustering-guided Mamba module (the bottom figure in \autoref{fig1}), where the spectral Mamba block and the clustering-guided spatial Mamba block are sequentially performed to learn the spatial-spectral information in HSI. The clusters are achieved via the residual matrix \( \mathcal{H}^{\text{res}}_{l\mathcal{M}}=\mathcal{P}_{\mathcal{M}^{res}}(\mathcal{H}_{l}^{\text{res}
    })\).

    \item \textbf{Implementing Residual Matrix as Clustering Maps}: In \autoref{fig1}, the residual matrix \( \mathcal{H}^{\text{res}}_{l\mathcal{M}} \), highlighted by the red bounding box  2, is used to estimate the clustering maps. We found that there is a clear spatial clustering effect in the doubly stochastic matrix, \( \mathcal{H}^{\text{res}}_{l\mathcal{M}} \in \mathbb{R}^{L \times n \times n}\) where $L=H \times W$, $H$ is the image hight, and $W$ is the image width and $n$ is the number of streams/lanes. Each element in \( \mathcal{H}^{\text{res}}_{l\mathcal{M}} (i, j) \in \mathbb{R}^{H \times W}, 1 \leq i \leq n,  1 \leq j \leq n\) is the soft class membership map of one cluster. In total we have $n \times n$ clusters. 
    \item \textbf{Electromagnetic spectrum–aware residual stream}: As shown by the red bounding box 3 in \autoref{fig1}, to generate multiple streams, instead of duplicating the input data, we split the input data into four physical meaningful spectral groups, i.e., FULL, VIS, NIR, SWIR1, SWIR2, leading to a total of 5 (\(n = 5\)) streams. 
\end{itemize}




\subsection{Clustering-Guided Mamba}

Cluster-guided Mamba (CGM) serves as the feature extraction function \(F(\cdot)\), shown in red bounding box 1 in \autoref{fig1}, which contains spectral Mamba and spatial cluster Mamba that are combined sequentially.

\subsubsection{Spectral Mamba}
As shown in the left dash bounding box in \autoref{fig1}, the input feature \(\mathcal{H}^{\text{pre}}_l \boldsymbol{R}_l \in \mathbb{R}^{B \times L \times D}\) is split into $G$ groups along the channel dimension, forming a total of \(G\) tokens with each token owning $M$ spectral channels, and \(D = GM\). These spectral tokens are processed by the Mamba algorithm. 
\begin{align}
    \begin{aligned}
        \overline{\mathcal{H}^{\text{pre}}_l \boldsymbol{R}_l}^{\text{spe}} &= \mathcal{H}^{\text{pre}}_l \boldsymbol{R}_l \\ &+ \text{Reshape} (\textbf{Mamba}(\text{ChannelSplit}(\mathcal{H}^{\text{pre}}_l \boldsymbol{R}_l))
    \end{aligned}
\end{align}
The \(\overline{\mathcal{H}^{\text{pre}}_l \boldsymbol{R}_l}^{\text{spe}}\) is the output of spectral Mamba. Then, it will be fed into Cluster-guided spatial Mamba (CGSM) for spatial feature learning.

\subsubsection{Spatial Cluster Mamba}

The right dash bounding box in \autoref{fig1} gives the overview of the CGSM Scanning. As said in \autoref{overall}, total \(n^2\) cluster maps are used for token selection in CGSM. By selecting the Top-k tokens in each spatial cluster matrix with shape \(H \times W\), \(n^2\) parallel spatial Mamba blocks are used to extract the spatial information for each element $(i,j)$ of $\mathcal{H}^{\text{res}}$, which can be expressed as the following:
\begin{align}
    \begin{aligned}
        \mathcal{T}^{i,j} &= \operatorname{Top\text{-}k}\!\left(\overline{\mathcal{H}^{\text{pre}}_l \boldsymbol{R}_l}^{\text{spe}}, \mathcal{H}^{\text{res}}_{:,i,j}\right), \\
        \hat{\mathcal{T}}^{i,j} &= \operatorname{CGSM}_{i,j}\!\left(\mathcal{T}^{i,j}\right),
        \quad i,j \in \{1,\dots,n\},\\
    \end{aligned}
\end{align}
where $\operatorname{CGSM}_{i,j}$ denotes a cluster-wise Spatial Mamba applied in parallel across all $(i,j)$ components. \(\mathcal{T}^{i,j}\) is the selected spatial tokens, while the \(\hat{\mathcal{T}}^{i,j}\) are the processed ones. After the parallel CGSM blocks are all finished, all the tokens are remapped to the original location in feature \(\boldsymbol{R}_l\), as follows:
\begin{align}
    \begin{aligned}
        \text{CGM}_{l}(\mathcal{H}^{\text{pre}}_{l} \boldsymbol{R}_{l}) &= \boldsymbol{R}_{l} + \textbf{Map}((\operatorname{sort}^{-1}(\hat{\mathcal{T}}^{i,j}))
    \end{aligned}
\end{align}
\(\operatorname{sort}^{-1}\) represents to recover to the original order. \textbf{Map} means put the processed feature at the original spatial location.

\subsection{Implementing Residual Matrix as Clustering Maps}

The pixel-based and object-based image analysis are two mainstream methods for remote sensing imagery analysis. One recent research study about the Sentinel-2 land use and land cover utilizes the superpixel-based object-level approach to define the token in the Mamba model, which reduces model parameters and can also increase the classification accuracy \cite{dewis2025multitaskglocalobiamambasentinel2}. Superpixels are also used in hyperspectral unmixing studies \cite{shi2022deep}. \cite{ahmad2025graphmamba} also studied the permutation and connectivity of tokens in Mamba, but it is not a cluster-wise method, leading to limited consideration of spatial consistency.

In this paper, we found that in the residual stream mixing matrix \(\mathcal{H}^{\text{res}}\), there are clear and consistent clustering phenomena that will guide the token selection in the Mamba model, contributing to fewer but more related tokens. Additionally, this clustering phenomenon leads us to analyze the hidden connection between different clusters and the semantic label in the ground truth. More detailed discussions are in \autoref{Clustering effect}.

\subsection{Physical Meaningful Multi-Stream Representations}
Instead of duplicating the input feature to build multi-stream representations HC and \textit{m}HC did, we fully consider the unique characteristic of the HSI cube from the electromagnetic spectrum perspective, by splitting the original HSI cube into non-overlapping sub-cubes i.e., visible light (VIS, 400-700 nm), near-infrared (NIR, 700-1000 nm), shortwave infrared 1 (SWIR1, 1000-1800 nm), shortwave infrared 2 (SWIR2, 1800-2500 nm), to expand the width of the neural network's input feature, hence increasing the dense connectivity and multi-path structure. For original HSI cube \(\mathbf{H} \in \mathbb{R}^{H \times W \times C}\), total \(n = 5\) parallel streams are defined as follows:
\begin{align}
    \begin{aligned}
        \mathbf{H} \in \mathbb{R}^{H \times W \times C}, \mathbf{H} = \left[\begin{matrix}
            \mathbf{h}_{\text{FULL}} ,
            \mathbf{h}_{\text{VIS}} ,
            \mathbf{h}_{\text{NIR}} ,
            \mathbf{h}_{\text{SWIR1}} ,
            \mathbf{h}_{\text{SWIR2}} 
        \end{matrix}\right]
    \end{aligned}
\end{align}
where \(\mathbf{h}_{i} \in \mathbb{R}^{H \times W \times C_e}\), \(C_e\) varies from the range of electromagnetic spectrum, \(e \in \{\text{FULL}, \text{VIS}, \text{NIR}, \text{SWIR1}, \text{SWIR2}\}\), \(\sum_{e=1}^{n} C_e = C\). All five streams are processed by the parallel embedding layers, with position embedding added to the "FULL" stream, forming stream representation \(\boldsymbol{R} \in  \mathbb{R}^{L \times n \times D}\), as denoted in algorithm \autoref{algo1}. This operation will maintain and enhances stability and scalability due to the manifold constraint of residual stream mixing matrix \(\mathcal{H}^{\text{res}}\), as well as enhance the interpretability of \(\mathcal{H}^{\text{res}}\), because \(\mathcal{H}^{\text{res}}\) reflects the mixing and interaction among \(n\) streams. An example is shown in \autoref{fig1}. We can see that the spatial pattern and structure are well preserved, but the intensity reflected by the reflectance of different spectral ranges is different.

\begin{algorithm}[htbp]
\caption{Our proposed model}
    \begin{algorithmic}[1] \label{algo1}
    \scriptsize
        \REQUIRE Hyperspectral image cube \(\mathbf{H} \in \mathbb{R}^{H \times W \times C}\), training samples mask set \(\mathcal{D} \in \mathbb{R}^{H \times W}\), four defined electromagnetic spectrum–aware residual stream with additional full bands \(\mathbf{E} \in \{\text{FULL}, \text{VIS}, \text{NIR}, \text{SWIR1}, \text{SWIR2}\}\) in \autoref{method}. Hidden dimension \(D\). Expansion rate \(n = 1+4=5\). Total \(L\) layers in \textit{m}HC blocks set \(\mathcal{L}\)
         \STATE initialize a residual stream list \(\mathcal{RS}\)
         \FOR{\(e\) \text{in} \(\mathbf{E}\)}
         \STATE get the corresponding cube \(\mathbf{H}_e \in \mathbf{R}^{H \times W \times C_e}\) for physical band \(e\)
         \STATE \(\mathbf{F}_{C_e}\) = \textbf{Embedding}(\(\mathbf{H}_e\))
         \IF {\(e = \text{FULL}\)}
         \STATE \(\mathbf{F}_{C_e} = \mathbf{F}_{C_e} + \text{Spatial Position Encoding}\) 
         \ENDIF
        \STATE append \(\mathbf{F}_{C_e}\) to \(\mathcal{RS}\)
        \ENDFOR
        \STATE get the residual stream \(\boldsymbol{R} \in \mathbb{R}^{L \times n \times D}\) by stacking \(\mathcal{RS}\) together 
        \FOR{\textit{m}HC layer \(l\) in block sets \(\mathcal{L}\)}
        \STATE \(\tilde{\boldsymbol{R}_{l}}\) = \text{RMSNorm} (\(\boldsymbol{R}_{l}\))
        \STATE \(\tilde{\mathcal{H}}_l^{\text{pre}} = \alpha_{l}^{\text{pre}} \cdot \text{tanh}(\theta_{l}^{\text{pre}} \tilde{\boldsymbol{R}}_{l}^T) + \mathbf{b}_{l}^{\text{pre}}\) 
        \STATE \(\tilde{\mathcal{H}}_{l}^{\text{post}} = \alpha_{l}^{\text{post}} \cdot \text{tanh} (\theta_{l}^{\text{post}}\tilde{\boldsymbol{R}}_{l}^T) + \mathbf{b}_{l}^{\text{post}}\)
        \STATE \(\tilde{\mathcal{H}}_{l}^{\text{res}} = \alpha_{l}^{\text{res}} \cdot \text{tanh} (\theta_{l}^{\text{res}}\tilde{\boldsymbol{R}}_{l}^T) + \mathbf{b}_{l}^{\text{res}}\)
        \STATE \(\mathcal{H}^{\text{pre}}_{l} = \sigma (\tilde{\mathcal{H}}^{\text{pre}}_{l})\)
        \STATE \(\mathcal{H}^{\text{post}}_{l} = 2\sigma (\tilde{\mathcal{H}}^{\text{post}}_{l})\)
        \STATE \(\mathcal{H}^{\text{res}}_{l} = \text{Sinkhorn-Knopp} (\tilde{\mathcal{H}}^{\text{res}}_{l})\)
        \STATE \(\hat{\boldsymbol{R}_{l}} = \mathcal{H}_{l}^{\text{res}} \boldsymbol{R}_{l} + \mathcal{H}_{l}^{\text{post}\top} (\text{CGM}_{l}(\mathcal{H}^{\text{pre}}_{l} \boldsymbol{R}_{l}))\)
        \STATE \(\bar{\boldsymbol{R}_{l}} = \text{RMSNorm}(\hat{\boldsymbol{R}_{l}})\)
        \STATE \(\bar{\mathcal{H}}_l^{\text{pre}} = \alpha_{l}^{\text{pre}} \cdot \text{tanh}(\theta_{l}^{\text{pre}} \bar{\boldsymbol{R}}_{l}^T) + \mathbf{b}_{l}^{\text{pre}}\) 
        \STATE \(\bar{\mathcal{H}}_{l}^{\text{post}} = \alpha_{l}^{\text{post}} \cdot \text{tanh} (\theta_{l}^{\text{post}}\bar{\boldsymbol{R}}_{l}^T) + \mathbf{b}_{l}^{\text{post}}\)
        \STATE \(\bar{\mathcal{H}}_{l}^{\text{res}} = \alpha_{l}^{\text{res}} \cdot \text{tanh} (\theta_{l}^{\text{res}}\bar{\boldsymbol{R}}_{l}^T) + \mathbf{b}_{l}^{\text{res}}\)
        \STATE \(\mathcal{H}^{\text{pre}}_{l} = \sigma (\bar{\mathcal{H}}^{\text{pre}}_{l})\)
        \STATE \(\mathcal{H}^{\text{post}}_{l} = 2\sigma (\bar{\mathcal{H}}^{\text{post}}_{l})\)
        \STATE \(\mathcal{H}^{\text{res}}_{l} = \text{Sinkhorn-Knopp} (\bar{\mathcal{H}}^{\text{res}}_{l})\) 
        \STATE \({\boldsymbol{R}_{l+1}} = \mathcal{H}_{l}^{\text{res}} \hat{\boldsymbol{R}}_{l} + \mathcal{H}_{l}^{\text{post}\top} (\text{FFN}_{l}(\mathcal{H}^{\text{pre}}_{l} \hat{\boldsymbol{R}}_{l}))\)
        \ENDFOR
        \STATE final feature \(h = \text{Mean}(\boldsymbol{R}_{L}, \text{dim}=2) \in \mathbb{R}^{L \times D}\)
        \STATE run a classification head on \(h\) to get the prediction logits
        \STATE calculate the cross-entropy using training samples
        \STATE update model parameters and repeat
    \end{algorithmic}
\end{algorithm}

\section{Experiments} \label{exp}

\begin{table*}[!h]
\caption{Quantitative performance of different classification methods in terms of OA, AA, $k$, as well as the accuracies for each class on the Indian Pines dataset with 10 \% training samples. The best results are in bold and colored shadow.}
\resizebox{\textwidth}{!}{
\begin{tabular}{ccc|cc|c|ccc|cc|c}
\hline
\multicolumn{1}{c|}{\multirow{2}{*}{Class No.}} & \multicolumn{1}{c|}{\multirow{2}{*}{Train Number}} & \multirow{2}{*}{Test Number} & \multicolumn{2}{c|}{CNN-based} & GAN-based & \multicolumn{3}{c|}{Transformer-based} & \multicolumn{2}{c|}{Mamba-based} &  \textit{m}HC-based                      \\ \cline{4-12} 
\multicolumn{1}{c|}{}                           & \multicolumn{1}{c|}{}                              &                              & SSRN      & SS-ConvNeXt        & MTGAN     & SSFTT  & SSTN         & GSC-ViT        & MambaHSI      & 3DSS-Mamba & \textbf{Ours}  \\ \hline
\multicolumn{1}{c|}{1}                          & \multicolumn{1}{c|}{5}                             & 37                           & 90.98     & 84.21              & 84.63     & 94.18  & 94.44        & 94.59          & 94.74         & 95.12      & \cellcolor[RGB]{251, 228, 213}\textbf{100.00}           \\
\multicolumn{1}{c|}{2}                          & \multicolumn{1}{c|}{143}                           & 1142                         & 97.54     & \cellcolor[RGB]{251, 228, 213}\textbf{99.53}     & 97.20      & 94.9   & 97.99        & 95.19          & 98.08                 & 93.22      & 99.13          \\
\multicolumn{1}{c|}{3}                          & \multicolumn{1}{c|}{83}                            & 664                          & 96.88     & 96.62              & 95.9      & 92.16  & 95.48        & 96.54          & 96.36               & 93.44      & \cellcolor[RGB]{251, 228, 213}\textbf{98.24}  \\
\multicolumn{1}{c|}{4}                          & \multicolumn{1}{c|}{24}                            & 189                          & 97.7      & 95.24              & 96.85     & 94.8   & 93.12        & \cellcolor[RGB]{251, 228, 213}\textbf{99.47} & 90.05                & 96.71      & \cellcolor[RGB]{251, 228, 213}\textbf{100.00}   \\
\multicolumn{1}{c|}{5}                          & \multicolumn{1}{c|}{48}                            & 387                          & 95.13     & 97.91              & 95.61     & 95.33  & \cellcolor[RGB]{251, 228, 213}\textbf{99.22}        & 98.45          & 97.67       & 96.71      & 95.82          \\
\multicolumn{1}{c|}{6}                          & \multicolumn{1}{c|}{73}                            & 584                          & 99.25     & 99.85              & 98.48     & 96.66  & 99.66        & 98.29          & 99.49       & 95.40      & \cellcolor[RGB]{251, 228, 213}\textbf{100.00}          \\
\multicolumn{1}{c|}{7}                          & \multicolumn{1}{c|}{3}                             & 22                           & 76.4      & \cellcolor[RGB]{251, 228, 213}\textbf{100.00}                & 12.00        & 84.92  & 68.18        & 77.27          & \cellcolor[RGB]{251, 228, 213}\textbf{100.00}    & 98.63      & 91.67   \\
\multicolumn{1}{c|}{8}                          & \multicolumn{1}{c|}{48}                            & 382                          & 99.53     & 99.76              & 99.95     & 99.62  & \cellcolor[RGB]{251, 228, 213}\textbf{100.00}          & \cellcolor[RGB]{251, 228, 213}\textbf{100.00}            & \cellcolor[RGB]{251, 228, 213}\textbf{100.00}   & 96.97      & \cellcolor[RGB]{251, 228, 213}\textbf{100.00}   \\
\multicolumn{1}{c|}{9}                          & \multicolumn{1}{c|}{2}                             & 16                           & 55.29     & 92.86              & 61.76     & 77.83  & 78.57        & 93.75          & \cellcolor[RGB]{251, 228, 213}\textbf{100.00}           & 72.22      & \cellcolor[RGB]{251, 228, 213}\textbf{100.00}   \\
\multicolumn{1}{c|}{10}                         & \multicolumn{1}{c|}{97}                            & 778                          & 96.19     & 97.81              & 95.43     & 92.21  & \cellcolor[RGB]{251, 228, 213}\textbf{98.97}        & 97.17          & 97.04   & 92.80      & 97.34 \\
\multicolumn{1}{c|}{11}                         & \multicolumn{1}{c|}{245}                           & 1965                         & 98.29     & 88.78              & 98.44     & 97.52  & 97.91        & 99.75          &\cellcolor[RGB]{251, 228, 213}\textbf{ 99.29 }               & 96.92      & 99.13 \\
\multicolumn{1}{c|}{12}                         & \multicolumn{1}{c|}{59}                            & 475                          & 97.97     & 97.92              & 95.10     & 90.48  & \cellcolor[RGB]{251, 228, 213}\textbf{100.00} & 95.16          & 99.16              & 90.64      & 96.79          \\
\multicolumn{1}{c|}{13}                         & \multicolumn{1}{c|}{20}                            & 164                          & 99.68     & \cellcolor[RGB]{251, 228, 213}\textbf{100.00}       & 98.76     & 96.59  & 99.39        & \cellcolor[RGB]{251, 228, 213}\textbf{100.00}   & 99.39                & 95.14      & \cellcolor[RGB]{251, 228, 213}\textbf{100.00}   \\
\multicolumn{1}{c|}{14}                         & \multicolumn{1}{c|}{126}                           & 1012                         & 99.57     & \cellcolor[RGB]{251, 228, 213}\textbf{99.91}              & 99.12     & 98.67  & 99.8         & 99.7           & 99.11              & 99.65      & \cellcolor[RGB]{251, 228, 213}\textbf{99.91}   \\
\multicolumn{1}{c|}{15}                         & \multicolumn{1}{c|}{39}                            & 308                          & 98.01     & 99.71              & 98.10     & 83.48  & 99.35        & 96.44          & \cellcolor[RGB]{251, 228, 213}\textbf{100.00}     & 95.68      & \cellcolor[RGB]{251, 228, 213}\textbf{100.00}   \\
\multicolumn{1}{c|}{16}                         & \multicolumn{1}{c|}{9}                             & 75                           & 97.38     & 97.53              & 92.50     & 89.93  & 89.33        & 97.3           & 85.33             & 85.71      & \cellcolor[RGB]{251, 228, 213}\textbf{98.79} \\ \hline
\multicolumn{3}{c|}{OA(\%)}                                                                                                         & 97.81     & 96.29              & 97.20     & 95.12  & 98.23        & 97.94          & 98.28              & 95.53      & \cellcolor[RGB]{251, 228, 213}\textbf{98.85} \\
\multicolumn{3}{c|}{AA(\%)}                                                                                                         & 93.49     & 96.73              & 88.80     & 92.46  & 94.49        & 96.19          & 97.23             & 94.90       & \cellcolor[RGB]{251, 228, 213}\textbf{98.55} \\
\multicolumn{3}{c|}{Kappa(\%)}                                                                                                      & 97.74     & 95.96              & 97.08     & 94.76  & 97.98        & 97.65          & \cellcolor[RGB]{251, 228, 213}\textbf{98.21}              & 93.42      & 97.44 \\ \hline
\end{tabular}
\label{table:IP}
}
\end{table*}

\subsection{Classification Results}

\autoref{table:IP} shows the numerical results achieved by different methods on the Indian Pines dataset. Our approach outperforms the other methods on all metrics. In particular, our approach \mbox{\textit{m}HC-HSI}, achieves much better results on average accuracy (AA) and overall accuracy (OA), indicating that the proposed approach outperforms the other approaches in terms of preserving and classifying the small classes.

\begin{figure}[!h]
    \centering
    \includegraphics[width=0.49\textwidth]{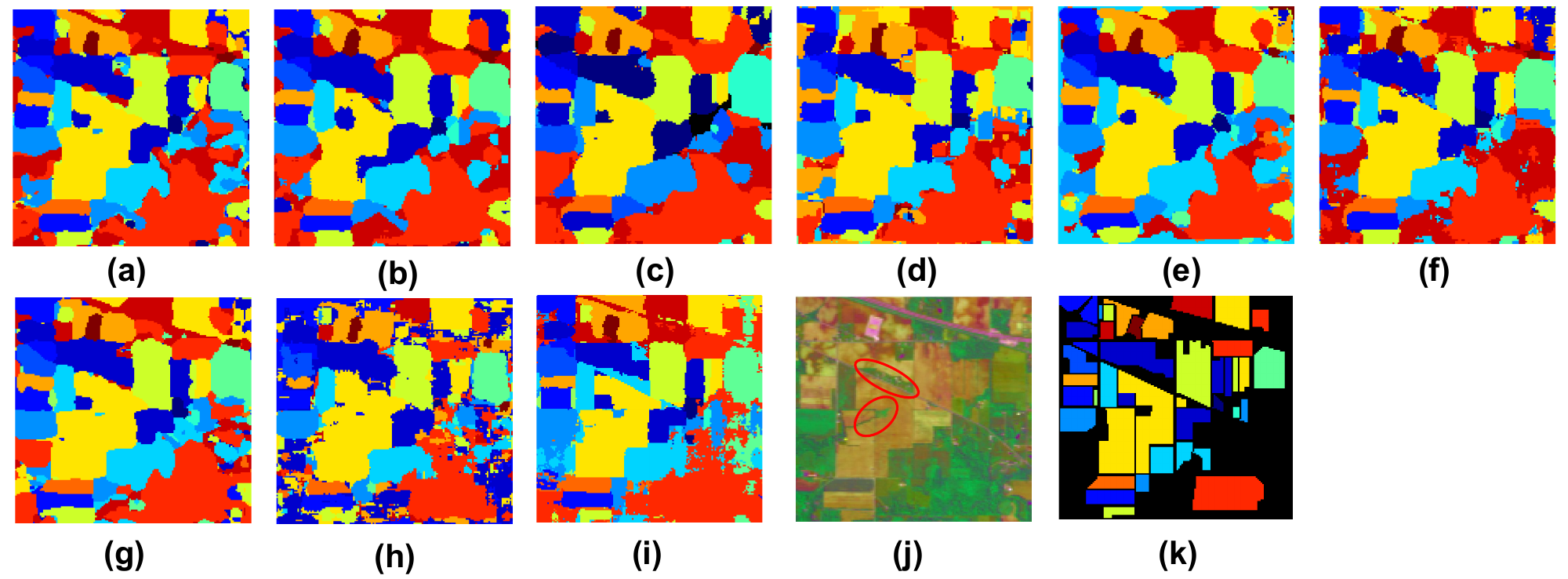}\\
    \includegraphics[width=0.49\textwidth]{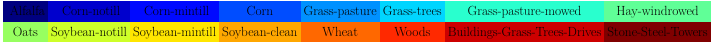}
    \caption{The Indian Pines classification map generated by different methods. (a) SSRN (b) SS-ConvNeXt (c) MTGAN (d) SSFTT (e) SSTN (f) GSC-ViT (g) MambaHSI (h) 3DSS-Mamba (i) \textit{m}HC-HSI (j) False Color Image (k) Ground Truth. Some red circles are shown on the  RGB image to illustrate the boundary preservation of our proposed model.}
    \label{maps}
\end{figure}

Additionally, as shown in \autoref{maps}, the proposed approach achieves a map that is not only the most consistent with the classification map, but also better at delineating the boundaries and small classes, as shown in the two red circle regions of interest areas.
\subsection{Impact on the expansion rate}

\begin{table}[!h]
\caption{Impact on the expansion rate \(n\).}
\resizebox{0.49\textwidth}{!}{
\begin{tabular}{c|cc}
\hline
Metrics & Expansion rate \(n\)=2 & Expansion rate \(n\)=4 \\ \hline

OA        & 98.95              & 98.54              \\
AA        & 97.65              & 97.77              \\
Kappa     & 95.98              & 95.80              \\ \hline
\end{tabular}
}
\label{expansion_rate}
\end{table}
Furthermore, we explore the impact on the expansion rate and visualize the key matrix, \(H^{\text{res}}\). Note that this experiment is under the setting of duplicating the input feature, i.e., duplicating the original HSI cube for \(n\) times as \textit{m}HC did, instead of being split into more physically meaningful spectrum bands. The numerical results are shown in \autoref{expansion_rate}, which demonstrate that the comparable classification performance is achieved when only duplicating the input HSI cube, but using the electromagnetic spectrum–aware residual stream design method, the results are better than the original duplication for the network's input width expansion. From the multi-view learning perspective, our proposed \textit{m}HC-HSI has more clear physical meaning, which can help us to understand the model inference.

\subsection{Clustering effect in \(\mathcal{H}^{\text{res}}\)} \label{Clustering effect}

As illustrated in \textit{m}HC paper \cite{xie2026mhc}, \(\mathcal{H}^{\text{pre}}\) serves as the role of the feature compression, by compressing the expanded \(n\) features into one representative feature. While, \(\mathcal{H}^{\text{post}}\) serves as feature reconstruction matrix to map the compressed feature to the original size. \(\mathcal{H}^{\text{res}}\) is the learnable mapping that mixes features within the \(n\) residual streams. More importantly, these three matrices are learned from data, which means they are feature-dependent parameters, see Algorithm~\autoref{algo1} and \textit{m}HC paper~\cite{xie2026mhc}.



\begin{figure}[!h]
    \centering
    \includegraphics[width=\linewidth]{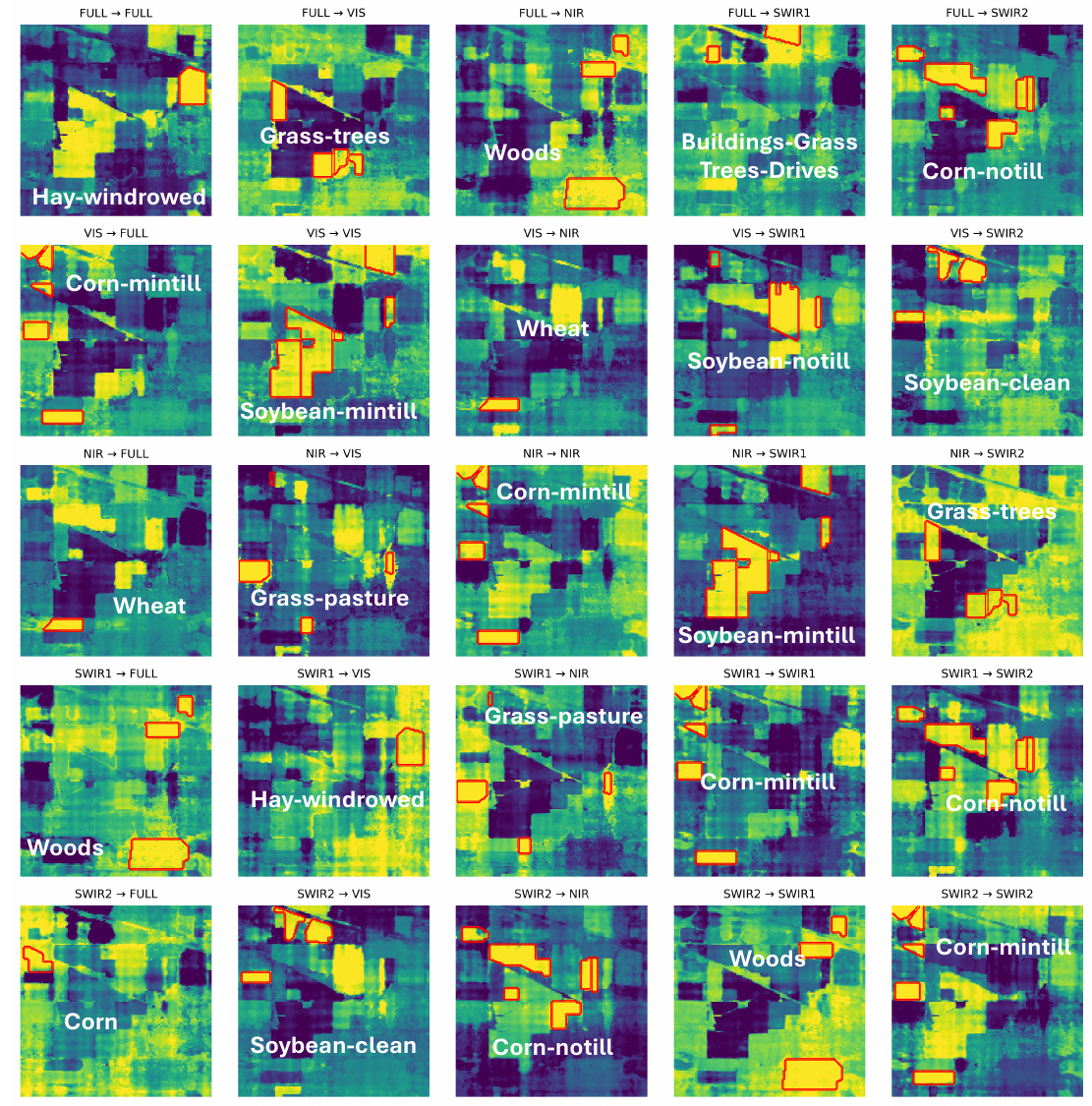}
    \caption{Visualization of the interpretable \(\mathcal{H}^{\text{res}}\) with overlaid class boundaries. The boundary is selected based on the highest mean value for each class boundary. White text shows the name of the category.}
    \label{H_res}
\end{figure}
\autoref{H_res} shows the correspondence between high-value regions in \(\mathcal{H}^{\text{res}}\) and category boundary.
These results are based on the \textit{m}HC-HSI. As we can see from this figure, each learnable mapping matrix, a total of 25 matrices, has its own unique category tendency.
Additionally, the bi-directional stream flow is asymmetric (off-diagonal element).
For example, ``FULL \(\rightarrow\) VIS" flow shows the high-value area in the Grass-trees (class 6), while ``VIS \(\rightarrow\) FULL" shows high value in the Corn-mintill area.
``Corn-notill, Corn-mintill, Corn" are more related to the SWIR bands, as they show higher values in \mbox{``FULL \(\rightarrow\) SWIR2"}, ``SWIR1 \(\rightarrow\) SWIR1", \mbox{``SWIR1 \(\rightarrow\) SWIR2"}, ``SWIR2 \(\rightarrow\) FULL", ``SWIR2 \(\rightarrow\) NIR", and ``SWIR2 \(\rightarrow\) SWIR2".
The ``notill" and ``mintill" mean crop residue covers the surface, or the mixture of soil and crop residue, which could lead to lower reflectance of \mbox{``Corn-notill''} and "Corn-mintill" at SWIR.

\section{Discussion and Conclusion} \label{discussion}

In this paper, we introduce—to the best of our knowledge— the first electromagnetic spectrum-aware residual stream splitting method, which replaces the standard \textit{m}HC residual design for hyperspectral image classification.
First, the clustering-guided Mamba module explicitly disentangles spatial and spectral feature learning. By capturing long-range context with linear complexity, it overcomes the correlation decay problems typically found in global sequence modeling. 
Second, the implementation of the residual matrix $\mathcal{H}^{\text{res}}$ as soft cluster membership maps allows the model to decompose the heterogeneous spatial patterns of HSI into smaller, manageable clusters. We observed clear clustering effects in $\mathcal{H}^{\text{res}}$ that align with physical ground-truth categories, providing a mechanistic insight into how hidden features are routed through the network.
Third, by dividing spectral bands into physically-meaningful groups, we leveraged domain-specific knowledge to ensure the ``streams" in $m$HC reflect real-world electromagnetic properties.
Experimental results on benchmark datasets confirm that our $m$HC-HSI model not only improves classification accuracy but also enhances model explainability.


\bibliographystyle{IEEEtran}
\bibliography{IEEEabrv,references}

\end{document}